\documentclass[pmlr,twocolumn]{jmlr}

\usepackage{amsmath}
\usepackage{booktabs}
 \usepackage{multirow}
\usepackage{longtable}% for long tables

\usepackage{booktabs}
\theorembodyfont{\upshape}
\theoremheaderfont{\scshape}
\theorempostheader{:}
\theoremsep{\newline}

\jmlrvolume{193}
\firstpageno{1}
\jmlryear{2022}
\jmlrworkshop{Machine Learning for Health (ML4H) 2022}

\title[On the effectiveness of a generative SSL approach to medical image segmentation]{Analysing the effectiveness of a generative model for semi-supervised medical image segmentation}% \titletag{\thanks{sample footnote}} 

\author{\Name{Margherita Rosnati}\Email{margherita.rosnati12@imperial.ac.uk}\\
\addr BioMedIA Group, Department of Computing, Imperial College London, UK
\AND
\Name{Fabio {De Sousa Ribeiro}}
\Email{f.de-sousa-ribeiro@imperial.ac.uk}\\
\addr BioMedIA Group, Department of Computing, Imperial College London, UK
\AND
\Name{Miguel Monteiro}
\Email{miguel.monteiro@imperial.ac.uk}\\
\addr BioMedIA Group, Department of Computing, Imperial College London, UK
\AND
\Name{Daniel Coelho {de Castro}} \Email{dacoelh@microsoft.com}\\
\addr Microsoft Research, UK\\
\addr BioMedIA Group, Department of Computing, Imperial College London, UK
\AND
\Name{Ben Glocker} \Email{b.glocker@imperial.ac.uk}\\
\addr BioMedIA Group, Department of Computing, Imperial College London, UK
}

\begin{document}

\maketitle

\begin{abstract}
Image segmentation is important in medical imaging, providing valuable, quantitative information for clinical decision-making in diagnosis, therapy, and intervention. 
The state-of-the-art in automated segmentation remains supervised learning, employing discriminative models such as U-Net.
However, training these models requires access to large amounts of manually labelled data which is often difficult to obtain in real medical applications. 
In such settings, semi-supervised learning (SSL) attempts to leverage the abundance of unlabelled data to obtain more robust and reliable models. 
Recently, generative models have been proposed for semantic segmentation, as they make an attractive choice for SSL. Their ability to capture the joint distribution over input images and output label maps provides a natural way to incorporate information from unlabelled images. 
This paper analyses whether deep generative models such as the SemanticGAN are truly viable alternatives to tackle challenging medical image segmentation problems. 
To that end, we thoroughly evaluate the segmentation performance, robustness, and potential subgroup disparities of discriminative and generative segmentation methods when applied to large-scale, publicly available chest X-ray datasets.
%\todo[inline]{Update downstream figures, ie run models. Update main text and conclusions. Update abstract. Add UNet experiments to appendix.}
\end{abstract}
\iffalse
\fi

\section{Introduction}
% DL in medical imaging is promising
Deep learning has shown promising results in medical image segmentation \citep{ronneberger2015u}. Applications include quantifying disease progression \citep{liu2020multi}, lesion volumes \citep{robben2020prediction}, tumour progression \citep{abdelazeem2020three} and radiotherapy planning \citep{oktay2020evaluation}.
% Limited 
However, large annotated datasets for training are scarce: despite data being routinely collected for all patients, expert annotations are prohibitively expensive and time-consuming to obtain in practice.
As a result, models are often trained on smaller datasets.
Consequently, models may not generalise well, with an often observed drop in performance when the data characteristics change---known as domain shift \citep{quinonero2008dataset}. 
In addition, potential biases in training data could cause deep learning methods to exacerbate health disparities \citep{obermeyer2019dissecting,adamson2018machine}. 

% SSL is a solution
Semi-supervised learning (SSL) \citep{chapelle2006semi} seeks to solve the problem of labelled data scarcity by leveraging information from unlabelled data in addition to the labelled data. 
% Discriminative models focus
Recent literature on SSL for segmentation focused on discriminative methods, mainly by augmenting the labelled training data through pseudo-labelling \citep{ke2020guided} or constraining models' latent representation by enforcing consistent predictions within similar datapoints \citep{ouali2020semi}. Methods based on these concepts have been extensively studied in the medical imaging context \citep{shurrab2022self, jiao2022learning}.
% Some generative stuff

Advances in SSL also span generative methods \citep{sajun2022survey}. The works by \citet{wei2018improving} and \citet{li2017triple} learn the distribution of the input data $\mathcal{X}$ to train a classifier more robustly. 
Recently, \citet{li2021semantic} expanded the modelling of $p(x)$ to the joint distribution $p(x,y)$ of images $x$ and their semantic segmentations $y$ for SSL. Their method showed promising preliminary results on generalisations to new tasks and datasets, characteristics particularly desirable in medical imaging. 
% To our knowledge, there is no in-depth analysis that evaluates the performance, generalisability, and robustness of joint image and labels generative SSL methods for medical images.

This work examines the aptitude of SSL deep generative models that learn the joint distribution of $p(x,y)$ to the challenging real-world medical image segmentation context.
We conduct an extensive empirical analysis on SemanticGAN \citep{li2021semantic} as a representative example, compared to a state-of-the-art fully supervised network and a semi-supervised discriminative ablation study on SemanticGAN for chest X-ray lung segmentation, to substantiate the following:
\begin{itemize}
    \item We provide thorough experimental evidence for SemanticGAN's generalisability for in- and out-of-domain tasks according to segmentation metrics and downstream disease classification performance;
    \item We uncover that the model's robustness cannot be explained by adversarial training alone; 
    \item We characterise the model's strengths and weaknesses regarding subgroup disparities and biases in the training data.
\end{itemize}

\section{Related work}
%Below, we first discuss semi-supervised learning for semantic segmentation, then we generative methods for semi-supervised learning, and then we describe semi-supervised methods in medical imaging. For more details on generic semi-supervised learning, we refer the reader to the relevant literature reviews~\citet{van2020survey,yang2021survey}. 
\paragraph{SSL for semantic segmentation}
% SSL in SS
Recent advances in SSL for semantic segmentation can be broadly categorised into two sometimes overlapping clusters: pseudo-label based and consistency constraint based. 
% Pseudo-labels
The former cluster use supervised learning architectures trained multiple times over pseudo-labels, which are labels produced by model predictions \citep{olsson2021classmix,chen2021semi,wang2022semi}. %ke2020guided,chen2020naive,
A limitation of these models is that they do not leverage the unlabelled datapoints for which their predictions are not confident.
% Consistency based 
The latter uses auxiliary tasks to generate a separable and information-dense latent representation of the data where similar datapoints are close \citep{ouali2020semi,liu2022perturbed}. For example, an idea borrowed from generative adversarial networks (GAN) is to use a discriminator to constrain the segmentation model to produce consistent segmentations \citep{hung2018adversarial}. These methods all suffer from poor generalisation to unseen data, and sometimes underperform fully supervised methods \citep{singh2008unlabeled,yang2021survey}.
\paragraph{Generative methods in SSL}
A stream of generative methods for SSL uses variational autoencoders (VAEs), modelling $p(x,y)$ through various latent variable models \citep{kingma2014semi,ehsan2017infinite,joy2020rethinking}. However, VAEs are rarely used in per-pixel classification.

In contrast, there are several GAN-inspired methods for generative SSL.
\citet{donahue2016adversarial} devised the ``bi-directional GAN" method,  where an encoder maps the image distribution to the GAN latent space. Further works \citep{dumoulin2016adversarially,kumar2017semi} use the latent space as the classification feature space. These works generate a compact representation of the data distribution of the images $p(x)$ and use it to predict the image label in a supervised manner.
Other authors employed both generative modelling and adversarial training to learn the image label. Using architectures derived from TripleGAN \citep{li2017triple}, \citet{dong2019margingan} and \citet{wu2019enhancing} propose to learn $p(x)$ through a GAN generator. Samples from the GAN are then concatenated to labels discriminatively generated by a classifier and fed to a discriminator for adversarial training. To the best of our knowledge, no GAN-based method other than \citet{li2021semantic} learns the joint distribution $p(x,y)$. For this reason, our work focuses on the latter.

\paragraph{SSL in medical imaging}
The literature on SSL methods in medical imaging broadly follows that of semantic segmentation, with pseudo-labelling methods \citep{han2022effective}, consistency regularisation methods \citep{basak2022embarrassingly} and adversarial learning \citep{peiris2021duo}. An in-depth review can be found at \citet{jiao2022learning}. Generative methods are seldom used. \citet{liu2019semi} used a GAN to reconstruct images as an auxiliary task for diabetic retinopathy screening, and \citet{zhang2022novel} used an architecture similar to TripleGAN to detect Parkinson’s Disease. 
%This work explores the potential of modelling the joint distribution of images and labels $p(x,y)$ in SSL for medical imaging by analysing a representative example's performance, generalisability and robustness.
To our knowledge, no previous work aims to assess methods that model the joint distribution of images and labels in SSL.
\section{Methods}
\iffalse
We compare a generative SSL method, SemanticGAN \citep{li2021semantic}, to a state-of-the-art fully supervsied semantic segmentation architecture, DeepLabV3 \citep{chen2017rethinking} and to an ablation study on SemanticGAN, where we replace the generative part of the architecture with the discriminative architecture of DeepLabV3. 
We choose SemanticGAN~\citet{li2021semantic} as the GAN-based method because it models the joint distribution of $p(x,y)$. The authors hypothesise that modelling the joint distribution yields superior robustness to changes in domain than discriminative methods, or methods only modelling $p(x)$. We aim to verify this hypothesis in a real-world medical imaging context. In addition, we propose an ablation study on the generative arm of SemanticGAN to determine whether adversarial training is sufficient for SemanticGAN's success. 
\fi
% START
The SemanticGAN~\citet{li2021semantic} authors hypothesised that modelling the joint distribution $p(x,y)$ of images and segmentations yields superior robustness to domain changes in comparison to discriminative methods, or methods only modelling $p(x)$. We aim to verify this hypothesis in a real-world medical imaging context. Specifically, we compare a generative SSL method, SemanticGAN \citep{li2021semantic}, to a state-of-the-art fully supervised semantic segmentation architecture, DeepLabV3 \citep{chen2017rethinking} and to an ablation study on SemanticGAN. Our proposed ablation study pertains to the generative arm of SemanticGAN to determine whether adversarial training is sufficient for SemanticGAN's success. 
% END

%Finally, we propose a second baseline designed to marry features of both SemanticGAN and DeepLabV3: an adversarially trained DeepLabV3. The latter can be considered an ablation study on SemanticGAN, testing the need for a generative model.
% In this section, we describe the architectures of the three methods.

\subsection{SemanticGAN}
SemanticGAN bases its architecture on StyleGAN2 \citep{karras2020analyzing}. Traditional GANs sample a noise vector $z\sim\mathcal{N}(0,I)$ and transform it through an upsampling architecture into an image. 
In contrast, StyleGAN2 transforms the noise vector into a higher dimensional style vector $w\in\mathcal{W}$ that is fed to the upsampling architecture at different stages, similarly to style transfer architectures. StyleGAN2 also borrows ideas from ResNet as it uses skip-connections to generate images of increasing resolution.
SemanticGAN modifies StyleGAN2 by using a second set of skip-connections to generate the image semantic segmentation, and adding a multi-scale patch-based discriminator \citep{wang2018high} for the segmentation. The method also adds an encoder from the image space into the $\mathcal{W}$ space \citep{richardson2021encoding}.

The model is trained against two discriminator models: the first, $D_r$, determines if the image is generated or real, and the second, $D_m$, determines if an image and segmentation pair is generated or real.
Following the original notation \citep{li2021semantic}, the objectives are defined as:
\newcommand{\realx}{x}  % previously x_r
\newcommand{\realy}{y}  % previously y_r
\newcommand{\fakex}{\tilde{x}}  % previously x_f
\newcommand{\fakey}{\tilde{y}}  % previously y_f
\begin{alignat}{3}
    \label{eq:semanticgan_gen_loss}
    &\mathcal{L}_G = &&\underset{\substack{z\sim p(z)\\(\fakex,\fakey)=G(z)}}{\mathbb{E}}&[&\log (1-D_r(\fakex)) \\ \nonumber
    &&&&&+ \log (1-D_m(\fakex, \fakey))] \,,  \\
    \label{eq:semanticgan_dr_loss}
    &\mathcal{L}_{D_r} = - &&\underset{\realx\sim\mathcal{D}_u}{\mathbb{E}}&[&\log D_r(\realx)] \\ \nonumber
    &&-& \underset{\substack{z\sim p(z)\\\fakex=G_X(z)}}{\mathbb{E}}&[&\log (1-D_r(\fakex))] ,  \\
%\end{alignat}
%\begin{alignat}{3}
    \label{eq:semanticgan_dm_loss}
    &\mathcal{L}_{D_m} =-&&\underset{(\realx, \realy)\sim\mathcal{D}_l}{\mathbb{E}}&[&\log D_m(\realx,\realy)] \\ \nonumber
    &&-& \underset{\substack{z\sim p(z)\\(\fakex,\fakey)=G(z)}}{\mathbb{E}}&[&\log (1-D_m(\fakex, \fakey))],  
\end{alignat}
where $G$ is the generator, $\mathcal{D}_u$ and $\mathcal{D}_l$ are the unlabelled and labelled datasets, respectively, and $(\realx,\realy)$ and $(\fakex, \fakey)$ are resp.\ real and generated samples.

% SemanticGAN encoder
Lastly, the model is also composed of an encoder $E$, which maps images to latent style representations compatible with StyleGAN2. The latter is trained with both an image reconstruction loss $\mathcal{L}_u$ and a semantic segmentation reconstruction loss $\mathcal{L}_s$:
\begin{alignat}{1}
\mathcal{L}_u = &  \underset{x\sim\mathcal{D}}{\mathbb{E}}[\mathcal{L}_{\text{LPIPS}}(x, G_X(E(x))) \\ \nonumber
&+\,\lambda_1\|x-G_X(E(x))\|_2^2] \,, \\
\label{eq:semanticgan_seg_loss}
\mathcal{L}_s = & \underset{(x, y)\sim\mathcal{D}_l}{\mathbb{E}}[\mathbf{H}(y, G_Y(E(x))) \\ \nonumber
&+\,\mathbf{DC}(y, G_Y(E(x)))] \,,     
\end{alignat}
where $\mathcal{L}_{\text{LPIPS}}(\cdot,\cdot)$ is the \textit{Learned Perceptual Image Patch Similarity} distance \citep{zhang2018unreasonable}, $\mathbf{H}$ is the cross entropy loss, $\mathbf{DC}$ is the Dice coefficient loss \citep{isensee2018nnu} and $\mathcal{D} = \mathcal{D}_l\cup \mathcal{D}_u$.
% SemanticGAN training
The model is trained in two steps. Firstly, the image and segmentation generative branch is trained along with the discriminators. Secondly, the weights of the generative branch are frozen, and the encoder is trained. 
We use the architectures in the original paper, where the modified StyleGAN2 generator has 31M parameters, the StyleGAN2 discriminator $D_r$  has 28M, the multi-scale patch-based discriminator $D_m$ has 8M, and the encoder $E$ has 7.4B parameters.

% Baselines
\subsection{Fully supervised baseline}
\label{Sup-UNet}
We compare SemanticGAN with DeepLabV3 on a 101-layer ResNet \citep{he2016deep} backbone, totalling 59M parameters. DeepLabV3 is considered state-of-the-art in semantic segmentation; its success based on augmenting the field of view of the network through atrous convolutions.
The architecture was optimised with a similar loss function to the SemanticGAN encoder $\mathcal{L}_s$ (Eq.~\eqref{eq:semanticgan_seg_loss}) for comparability:
\begin{alignat}{1}
\mathcal{L}_{\text{SupOnly}} = \underset{(x, y)\sim\mathcal{D}_l}{\mathbb{E}}[&\mathbf{H}(y, \text{DL}(x)) \\ \nonumber
&+\,\mathbf{DC}(y, \text{DL}(x))] \,,    
\end{alignat}
where DL stands for DeepLabV3. We call this supervised baseline SupOnly.
%\todo[inline]{"Sup-ResNet" and "Adv-ResNet"}
\subsection{Non-Generative Adversarial Network}
For the ablation study, we employ the architecture of DeepLabV3 to produce segmentation and train the model adversarially using SemanticGAN's architecture of $D_m$, which together have 67M parameters.
During training, we pass labelled or unlabelled images through the CNN architecture. We concatenate the resulting segmentation with the image and pass it through a discriminative model, charged with determining whether the pair comes from the labelled data distribution. 
In mathematical terms, the objective functions are (cf.\ Eqs.~\eqref{eq:semanticgan_gen_loss} and~\eqref{eq:semanticgan_dm_loss}):
\begin{alignat}{1}
&\mathcal{L}_{\text{Seg}} = \quad\underset{x\sim\mathcal{D}}{\mathbb{E}}
    [\log (1-D_m(x,\text{DL}(x)))] \,, \\
    &\mathcal{L}_{\text{Adv}} = \quad-\underset{(x, y)\sim\mathcal{D}_l}{\mathbb{E}}
    [\log D_m(x, y)] \\ \nonumber
    &\quad+ 
    \underset{x\sim\mathcal{D}}{\mathbb{E}}
    [\log (1-D_m(x,\text{DL}(x)))] \,,
\end{alignat}
where $\mathcal{L}_{\text{Seg}}$ optimises the weights of the discriminative segmentation model, and $\mathcal{L}_{\text{Adv}}$ optimises the weights of the adversarial model.
We call this adversarially trained method SemanticAN.

By design, SemanticAN is an ablation on SemanticGAN, testing whether the adversarial training is sufficient for an SSL model to generalise well, or whether learning the distribution of $p(x)$ is beneficial.

In Appendix~\ref{sec:Exp1-appendix}, we also compare SemanticGAN to SupOnly and SemanticAN based on a U-Net backbone \citep{ronneberger2015u} (578k param.) to evaluate the requirement of large models.

\section{Datasets}
% Choice of target
In order to compare and contrast the applicability of different approaches, we test them on the task of chest X-ray lung segmentation. Chest X-rays are one of the most common radiological examinations, and automatically extracted features from anatomical regions such as the lungs can support clinical decisions.
% dataset details - do we need more?

%Large open-source datasets of chest X-ray exist~\citet{wang2017chestx,irvin2019chexpert,shiraishi2000development}, many of which contain lungs segmentation~\citet{tang2019xlsor,van2006segmentation,jaeger2014two}, allowing us to test the model on different domains. 
We use the ChestX-ray8 \citep{wang2017chestx} (n=108k) and CheXpert \citep{irvin2019chexpert} (n=76k) unlabelled datasets and the NIH \citep{tang2019xlsor} (n=95), JSRT \citep{van2006segmentation} (n=431), and Montgomery \citep{jaeger2014two} (n=138) labelled datasets, where the NIH dataset is an annotated subset of the ChestX-ray8 dataset. A detailed description of the datasets can be found in Appendix table~\ref{tab:datasets}, while thumbnails of the datasets can be found in Appendix figures~\ref{fig:ds-thumb} and~\ref{fig:ds-thumb-2}, and raining details in Appendix~\ref{sec:training-deets}.

\section{Experiments}
\subsection{Model performance and in- and out-of-domain generalisation}
\label{methods-1}
%\subsection{Method}
We assess the resilience of generative segmentation networks to dataset changes by training SemanticGAN with the ChestX-ray8 unlabelled dataset and the JSRT labelled dataset, and testing the performance on the hold-out set of JSRT, the NIH dataset, and the Montgomery dataset. 
In this manner, we test the model on in-domain data for the labelled dataset, in-domain data for the unlabelled dataset, and out-of-domain data.
We calculate the Dice coefficient, precision, recall, average surface distance, and Hausdorff distance in pixels for each population.
The precision and recall highlight methods prone to over- or under-segmentation, whereas the average surface distance is sensitive to the distance between segmentations, and the Hausdorff distance is sensitive to far away outliers.
We compare the performance of SemanticGAN to that of SupOnly and SemanticAN, where SupOnly is only trained on the JSRT labelled dataset.
\begin{table*}[!t]
\centering
\caption{Models performance w.r.t.\ ground truth segmentations. Reported as mean $\pm$ standard deviation over the dataset.}
\label{exp1}
\setlength{\tabcolsep}{3pt}
\begin{tabular}{@{}lccccc@{}}
\toprule
%            & \textbf{Dice}                    & \textbf{Precision}               & \textbf{Recall}                  & \begin{tabular}[c]{@{}c@{}}\textbf{Avg. surface }\\\textbf{distance}\end{tabular} & \begin{tabular}[c]{@{}c@{}}\textbf{Hausdorff }\\\textbf{distance}\end{tabular}  \\
            & Dice                    & Precision               & Recall                  & \begin{tabular}[c]{@{}c@{}}Avg. surface\\distance\end{tabular} & \begin{tabular}[c]{@{}c@{}}Hausdorff\\distance\end{tabular}  \\
% \midrule
\midrule
&\multicolumn{5}{c}{JSRT (labelled in-domain)}                                                                                                                                                                                                                                     \\ 
\cmidrule{2-6}
%Sup-UNet    &     97.3 $\pm$ 1.1  & \textbf{98.4 $\pm$ 1.6}  & 96.2 $\pm$ 1.7  & 1.4 $\pm$ 1.8 & 5.5 $\pm$ 9.3\\
%Sup. DeepLab v3 & \textbf{99.1 $\pm$ 0.3} & \textbf{99.0 $\pm$ 0.4} & \textbf{99.3 $\pm$ 0.4} & \textbf{0.4 $\pm$ 0.2} & \textbf{1.2 $\pm$ 0.6}\\
SupOnly &\textbf{99.3 $\pm$ 0.4} & \textbf{99.1 $\pm$ 0.4} & \textbf{99.5 $\pm$ 0.5} & \textbf{0.3 $\pm$ 0.2} &\textbf{1.1 $\pm$ 0.4} \\
%Adv. DeepLab v3 & 94.8 $\pm$ 2.5 &         92.2 $\pm$ 4.3 &           97.6 $\pm$ 1.4 &         2.9 $\pm$ 1.83 &         13.0 $\pm$ 12.0 \\
SemanticAN &91.2 $\pm$ 5.2 &93.3 $\pm$ 3.7 &89.8 $\pm$ 7.6 &3.7 $\pm$ 2.0 & 14.7 $\pm$ 7.1 \\
SemanticGAN &     98.2 $\pm$ 1.3  &        98.3 $\pm$ 1.5  &          98.1 $\pm$ 1.2 &         0.7 $\pm$ 0.6 &          2.5 $\pm$ 2.4\\ 
\midrule
&\multicolumn{5}{c}{NIH (unlabelled in-domain)}                                                                                                                                                                                                                                    \\ 
\cmidrule{2-6}
%Sup-UNet        & 80.2 $\pm$ 15.9 & \textbf{91.0 $\pm$ 15.8} & 72.9 $\pm$ 17.2 & 8.6 $\pm$ 7.3 & 36.0 $\pm$ 24.4 \\
%Sup. DeepLab v3 & \textbf{90.6 $\pm$ 7.1} & \textbf{94.3 $\pm$ 6.4} & 87.6 $\pm$ 9.3 &            6.1 $\pm$ 5.71 &        29.0 $\pm$ 23.1 \\
SupOnly & \textbf{90.3 $\pm$ 8.1} & \textbf{95.3 $\pm$ 4.6} & 86.5 $\pm$ 11.4 &          \textbf{4.8 $\pm$ 4.8} &25.5 $\pm$ 22.7  \\
%Adv-UNet    & 81.6 $\pm$ 9.6  & 76.3 $\pm$ 12.6 & 88.8 $\pm$ 7.7  & 22.1 $\pm$ 7.7 & 71.1 $\pm$ 17.6 \\

%Adv. DeepLab v3 &73.9 $\pm$ 14.4 &          65.6 $\pm$ 18.4 &         87.4 $\pm$ 8.92 &          15.0 $\pm$ 7.15 &47.1 $\pm$ 17.2 \\
SemanticAN &72.4 $\pm$ 19.6 &75.4 $\pm$ 18.8 &71.2 $\pm$ 21.2 &10.9 $\pm$ 7.13 &35.5 $\pm$ 17.1 \\
SemanticGAN     & 90.1 $\pm$ 3.4 &          89.9 $\pm$ 6.5  &         \textbf{90.9 $\pm$ 5.0 } & 4.9 $\pm$ 3.3 & \textbf{18.2 $\pm$ 16.0}\\ 
\midrule
&\multicolumn{5}{c}{Montgomery (out-of-domain)}                                                                                                                                                                                                                                    \\ 
\cmidrule{2-6}
%Sup-UNet    & \textbf{92.7 $\pm$ 7.9}  & \textbf{96.6 $\pm$ 9.9}  & 89.4 $\pm$ 6.9  & \textbf{3.5 $\pm$ 4.7 } &\textbf{13.7 $\pm$ 17.8} \\
%Sup. DeepLab v3 & \textbf{96.5 $\pm$ 2.8} & \textbf{99.1 $\pm$ 1.2} & \textbf{94.1 $\pm$ 4.5} &          \textbf{1.3 $\pm$ 1.4} & \textbf{5.7 $\pm$ 8.5} \\
SupOnly &\textbf{96.5 $\pm$ 2.3} &\textbf{98.8 $\pm$ 0.7} & \textbf{94.4 $\pm$ 4.0} &\textbf{1.3 $\pm$ 0.7} & \textbf{5.5 $\pm$ 7.2} \\
%Adv-UNet    & 90.3 $\pm$ 6.4  & 87.0 $\pm$ 8.9  & \textbf{94.5 $\pm$ 4.0}  & 17.6 $\pm$ 8.5 & 72.3 $\pm$ 27.2 \\
%Adv. DeepLab v3 &  83.5 $\pm$ 22.0 &         82.3 $\pm$ 21.9 &        85.8 $\pm$ 21.6 &         10.8 $\pm$ 9.22 &44.4 $\pm$ 31.1 \\
SemanticAN &57.9 $\pm$ 33.4 &62.4 $\pm$ 31.0 &55.5 $\pm$ 35.1 &19.3 $\pm$ 15.1 &54.8 $\pm$ 38.9 \\
SemanticGAN &     91.3 $\pm$ 5.6  &         92.6 $\pm$ 7.9  &         90.4 $\pm$ 4.2  &         8.8 $\pm$ 10.5 & 34.2 $\pm$ 39.1        \\
\bottomrule
\end{tabular}
\end{table*}

The results summarised in Table~\ref{exp1} show SupOnly consistently outperformed SemanticGAN and SemanticAN when tested on JSRT and Montgomery. %(p-vals$<$.001). 
When tested on NIH, SemanticGAN outperformed SupOnly in terms of recall
% (p-val$<$.001)
and Hausdorff distance% (p-val$<$.05)
, whereas Dice score, precision and average surface distance were inconclusive when comparing the two.
These findings are in line with a stream of literature \citep{yang2021survey} %singh2008unlabeled
claiming that SSL methods do not consistently outperform fully supervised methods. 

Nevertheless, SemanticGAN produced Dice, precision and recall scores on average greater than 90\%, and an average surface distance of at most 8.8 pixels. 
The results indicate that the performance of SemanticGAN on datasets from a different distribution is consistent, although not consistently superior to baselines.

%\subsection{The power of image generation}
SemanticAN underperformed SemanticGAN and the supervised baseline% (p-vals$<$.001)
, whether presented with scans from the distribution of a labelled training set, unlabelled training set or a new dataset.
Its sub-par performance on NIH suggests that the model did not exploit the unlabelled dataset $\mathcal{D}_u$, and overfit to the training data distribution $\mathcal{D}_l$.
We find that the adversarial training alone is insufficient to learn the imaging data distribution and distorts the training signal, resulting in a weaker model than both SemanticGAN and the supervised baseline. 
\subsection{Downstream task performance for disease classification}
\label{methods-2}
\begin{figure*}[!t]
    \centering
    \includegraphics[width=\textwidth]{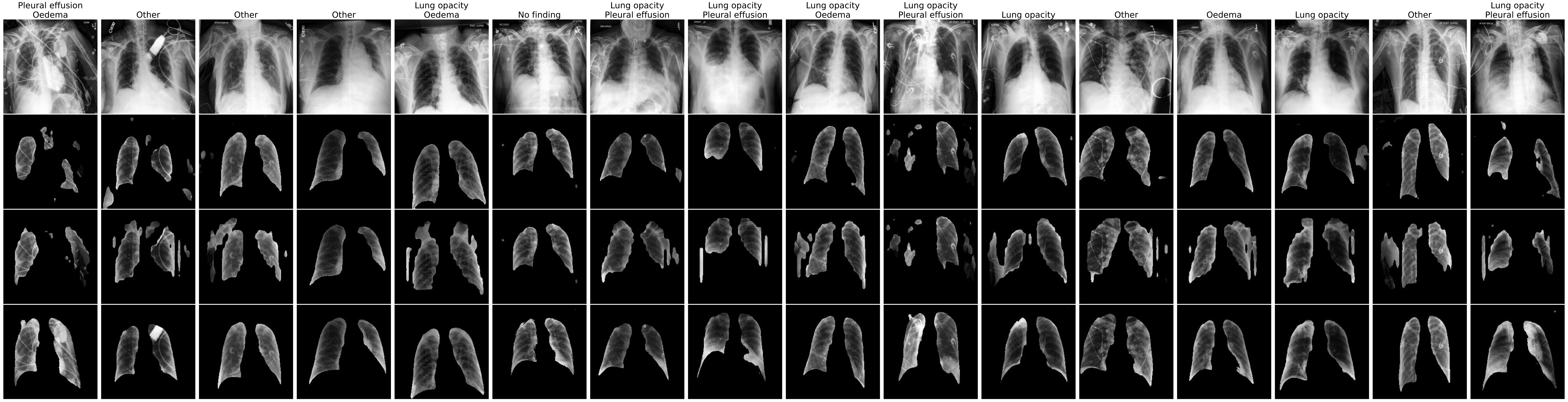}
    \caption{CheXpert images masked by models trained on ChestX-ray8 and JSRT. 
    First row: unmasked image; second row: SupOnly segmentation; third row: SemanticAN segmentation; fourth row: SemanticGAN segmentation. Titles: pathology label.}
    \label{fig:unsup_vis}
\end{figure*}
As a secondary indicator of the model performance, we evaluate the segmentation algorithm on a downstream task. 
This evaluation method benefits from not requiring ground truth segmentations, allowing us to test on larger datasets, such as CheXpert and ChestX-ray8.
We choose the disease classification downstream task, where we use the segmentation algorithm to mask the background to the lungs and detect different lung abnormalities. 

% For this purpose, we compare the performance of a classification model trained on unmasked images, images masked with SemanticGAN, and images masked with SupOnly and SemanticAN on the CheXpert dataset.
For this purpose, we compare the performance of a classification model trained on original unmasked images and images masked with SupOnly, Semantic AN, and SemanticGAN from the CheXpert dataset.
Unfortunately, we cannot compare these performances to that of the CNN trained on images segmented by clinicians due to the absence of such segmentations in the CheXpert dataset. The implementation details can be found in Appendix~\ref{sec:app-ex2}.
Within the reported 14 pathologies, we pick three clearly visible within the lung area and with a significant prevalence: pleural effusion (35.6\%), lung opacity (47.8\%) and oedema (18.9\%), where lung opacity and pleural effusion co-occur in 28.5\% of patients, lung opacity and oedema co-occur in 16.5\% of patients, pleural effusion and oedema co-occur in 15.6\% of patients, and all three pathologies co-occur in 9.5\% of patients.
We hypothesise that the information in the lung area is sufficient to classify the image correctly. Hence, a perfectly segmented image should be as predictive as the unmasked image and more predictive than any erroneously segmented one.
We train a binary classifier for each pathology and segmentation method and report the area under the receiver operator characteristic curve (AUROC) on the test images segmented with the same method as the training dataset. We provide confidence intervals by training the binary classifier with five different seeds.
In addition, we inspect a sample of the segmented images for a qualitative assessment.

Figure \ref{fig:unsup_vis} shows examples of patients' scans segmented with SemanticGAN and the two baselines. Both SupOnly's and SemanticAN's segmentations have irregular and disconnected shapes. They often mistake some low-intensity background areas for lungs, and under-segment high-intensity areas in the lungs, such as medical devices.
SemanticGAN's segmentations are anatomically plausible in shape across patients. However, the segmentation is sometimes misaligned with the lung -- for example on the last row of Figure \ref{fig:unsup_vis}, the top-left segmentation starts before the lung area.

\begin{table*}[!t]
\centering
\caption{Diagnosis classification AUROC for images masked using different segmentation models on the CheXpert dataset (out-of-domain). Reported as mean ±
standard deviation over the dataset.}
\label{tab:unsup}
\setlength{\tabcolsep}{5pt}
\begin{tabular}{@{}lccc@{}} 
\toprule
%               & \begin{tabular}[c]{@{}c@{}}Lung\\ opacity\end{tabular} 
%               & \begin{tabular}[c]{@{}c@{}}Pleural\\effusion \end{tabular} & Oedema  \\
& Opacity & Effusion & Oedema\ \\
\midrule
%\multicolumn{1}{c}{}& \multicolumn{3}{c}{CheXpert (out-of-domain)}           \\ 
%\cmidrule{2-4}
SupOnly         & 67.4 $\pm$ 0.2          & 82.0 $\pm$ 0.1             & 77.8 $\pm$ 0.2      \\
SemanticAN       & 66.6 $\pm$ 0.2        & 78.9 $\pm$ 0.2             & 77.2 $\pm$ 0.1      \\
SemanticGAN    & \textbf{67.9 $\pm$ 0.1} & \textbf{82.2 $\pm$ 0.2}    & \textbf{78.1 $\pm$ 0.2} \\
\midrule
%\begin{tabular}[c]{@{}c@{}}Unmasked\\ image\end{tabular} 
Unmasked & \textbf{70.1 $\pm$ 0.3}       & \textbf{85.0 $\pm$ 0.2}   & \textbf{79.1 $\pm$ 0.1}            \\
%Img            & NA & NA & 80.1 \\
\bottomrule
\end{tabular}
\end{table*}
Table~\ref{tab:unsup} shows the results of the classification tasks.
In contrast with the results in Section~\ref{methods-1}, SemanticGAN outperformed both SupOnly and SemanticAN for the classification of all tasks but Pleural Effusion, where SupOnly and SemanticGAN perform similarly. Paired with the qualitative assessment, we deduce that although SemanticGAN scores low on the segmentation metrics in Section~\ref{methods-1}, its segmentations are consistent in shape and may more robustly capture the anatomical region of interest for disease classification.

In addition, an investigation of how the pathologies' co-occurrence affect predictions found that all models predicted worse for patients with co-occurrences than for patients without co-occurrence. Our intuition is that the more pathologies are present in the image, the noisier the signal for each individual pathology, and the harder it can be for a given model to classify the (masked) images. In addition, the hierarchy of performance between SupOnly, SemanticAN, SemanticGAN and Unmasked shown in Table~\ref{tab:unsup} was observed for patient subgroups with multiple pathologies.

Finally, the superior classification performance of the classifier trained on unmasked images indicates that lung segmentations are imperfect for all methods, or other image regions may play an important role in the prediction of disease.
\subsection{Performance across subgroups and training bias}
\begin{figure*}[!t]

    \includegraphics[width=\textwidth]{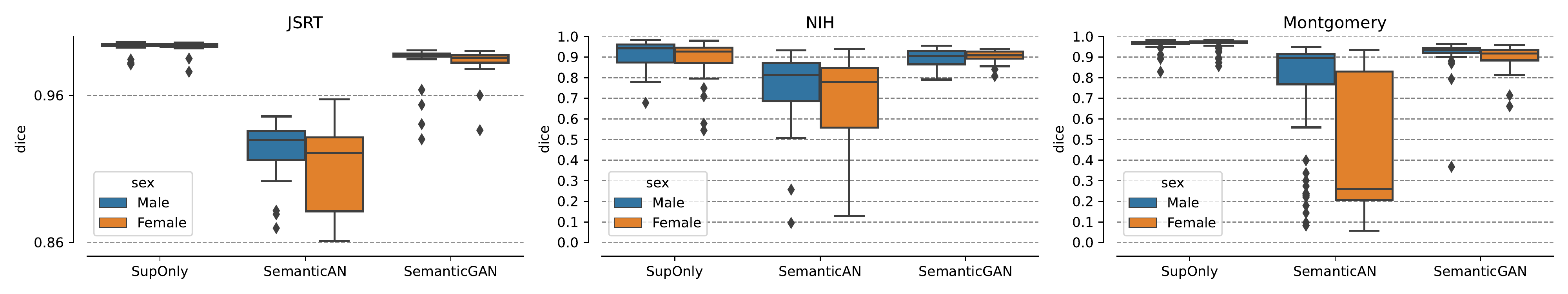}
 
 \caption{Stratification of segmentation results (Sections~\ref{methods-1}) by biological sex.}
 \label{fig:sup-sex}
\end{figure*}

\paragraph{Subgroup performance}
\label{subgroup-perf}
%\subsection{Method}
%We investigate potential biases in the model by stratifying the results of Sections~\ref{methods-1} and~\ref{methods-2} by biological sex \cite{larrazabal2020gender}.

Figure~\ref{fig:sup-sex} shows the stratification by the patient's biological sex of the Dice similarity coefficient derived in Section~\ref{methods-1}.
A corresponding numerical table can be found in Appendix Table~\ref{fig:sup-sex-numbers}.
% variance
%For JSRT and NIH, SemanticGAN achieved a lower variance in Dice scores than Sup-UNet and Adv-UNet. 
% within each model, females vs males
The relative model performance for females versus males was mixed for all models, except for SemanticAN.
The latter performed statistically significantly better for males than females for all datasets. 
On the other hand, SupOnly performed comparably for the two subgroups for all datasets, and SemanticGAN performed comparably for the subgroups for all datasets but Montgomery, where it performed better for males.
As SemanticAN is overfitting the labelled training data, it is unsurprising that it carries the same bias as on the training group on other groups. SemanticGAN showed to be more robust to biases than SemanticAN, yet worse than the fully supervised baseline.

When comparing models for each subgroup, %the models performed broadly in line with their performance in Section~\ref{methods-1}.
SemanticAN underperformed both SupOnly and SemanticGAN for all subgroups and datasets, and SupOnly outperformed SemanticGAN for all subgroups but females in NIH, where they performed comparably. %Given the results in Section~\ref{methods-1}, these results were in line with our expectations.
These results are consistent with the findings in Section~\ref{methods-1}.

A similar analysis stratifying the results of Section~\ref{methods-2} can be found in Appendix~\ref{sec:Exp3-appendix}.

\iffalse
%For each subgroup, SemanticGAN outperformed its baselines for all datasets but Montgomery, where Sup-UNet outperformed it for males. Sup-UNet performed better than Adv-UNet for both subgroups for JSRT and Montgomery. Otherwise, the models performed statistically equally to each other. 

Overall, the methods' relative performance was preserved for each patient subgroup. 
Any significant prediction discrepancy between subgroups observed on the baseline methods was not necessarily observed in SemanticGAN.
Prediction discrepancies between subgroups in other models did not necessarily translate into biases for SemanticGAN.
However, SemanticGAN's performance was sometimes more biased than its baselines, for example, in the case of predicting oedema. 
Overall, SemanticGAN was not consistently more or less biased than the other models.

\fi

\paragraph{Training Bias}
\label{training-bias}
A follow-up question to the subgroup stratification experiment is how stronger population biases in the training dataset affect the model performance over protected subgroups \citep{larrazabal2020gender}.
%In addition, we devise experiments to understand the impact of population biases in the training dataset. 
Specifically, we wish to learn whether the training of different parts of SemanticGAN with biased data affects the model performance; and whether the performance is more sensitive to a bias in either the labelled or the unlabelled datasets.

On that account, we generate a biased unlabelled dataset $\widetilde{\mathcal{D}_u}$ from ChestX-ray8, and a labelled biased dataset $\widetilde{\mathcal{D}_l}$ from JSRT, both composed of only the biological male, removing all biological females from the original datasets. 
We perform three sets of experiments, recapitulated in Appendix table~\ref{tab:detail-ex-bias}.
Firstly, we train a model solely on biased data, illustrating the most extreme circumstances. We call it the ``full bias" model.
Secondly, we train two models using the biased datasets at two different training phases. 
We train one model by training the generator $G$ with the biased datasets $\widetilde{\mathcal{D}}=\widetilde{\mathcal{D}_u}\cup\widetilde{\mathcal{D}_l}$ and then the encoder $E$ with the original datasets $\mathcal{D}$. We train a second model by training $G$ with the original datasets $\mathcal{D}$ and $E$ with the biased datasets $\widetilde{\mathcal{D}}$. We call these the ``biased generator" and ``biased encoder" models respectively.
Thirdly, we train one model with biased labelled data $\widetilde{\mathcal{D}_l}$ and original unlabelled data $\mathcal{D}_u$ for both $G$ and $E$, and a second model with original labelled data $\mathcal{D}_l$ and biased unlabelled data $\widetilde{\mathcal{D}_u}$. We call these the ``biased labelled dataset" and ``biased unlabelled dataset" models respectively.

For each iteration, we test the model on the same unbiased test datasets, JSRT, NIH and Montgomery, and compare their performance to the model from Sections~\ref{methods-1} and~\ref{methods-2}, trained on the original dataset $\mathcal{D}$---the ``control" model.

\begin{figure*}[!t]
    \centering
    \includegraphics[width=1\textwidth]{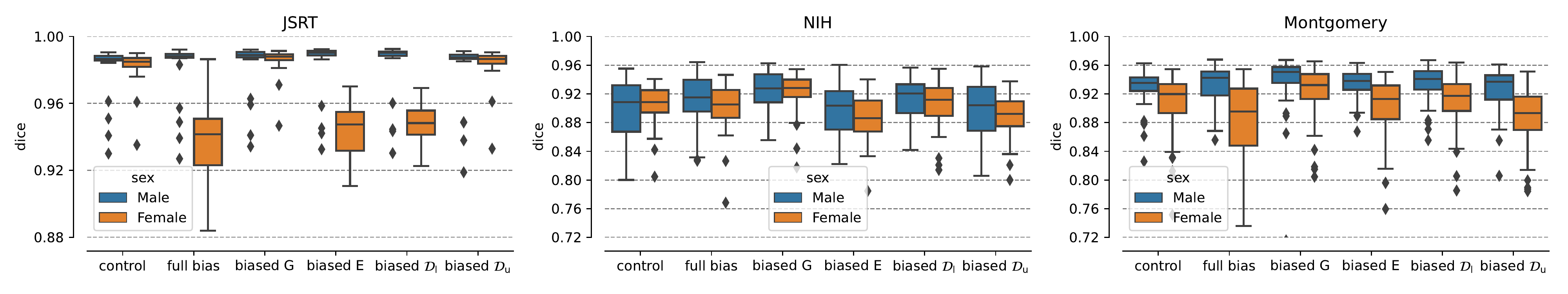}
    \caption{Impact of biological sex bias on model performance.}
    \label{fig:sex_bias}
\end{figure*}
Figure \ref{fig:sex_bias} presents the differences in Dice scores when training SemanticGAN with biases in different parts of the process. A detailed table with numerical values can be found in Appendix Table~\ref{fig:sex_bias_numerical}.
Overall, the training bias affected the performance of the JSRT dataset more than on the other datasets. 
For JSRT and biological females, the full bias, biased encoder and labelled data models performed significantly worse than the control. However, biasing generator and unlabelled data did not harm the model performance. In addition, the female bias did not affect the male subgroup performance.
However, for the NIH and Montgomery datasets, the discrepancy in model performance was smaller.
The control model performed better than the full bias and biased unlabelled dataset models for females, and the biased encoder model for females in NIH. 

The in-domain dataset results on JSRT for labelled data and NIH for unlabelled data were sensitive to biases in their training domain and the biased encoder. Intuitively, these results are coherent with the training paradigm, as the encoder explicitly minimises the difference between the datasets and the models' outputs.
The sensitivity to the biased unlabelled dataset in the out-of-domain dataset results also substantiates the hypothesis that SemanticGAN's generalisation properties derive from its training on unlabelled data. Consequently, if the model is presented with data significantly different from the unlabelled dataset, we expect SemanticGAN to decrease in performance.

\section{Conclusion}
In conclusion, this work provides a thorough analysis of the segmentation performance, robustness, and potential subgroup disparities of discriminative and generative segmentation methods when applied to large-scale, publicly available chest X-ray datasets. 

We found that SemanticGAN generates consistent predictions in shape (Section~\ref{methods-2}) and accuracy (Section~\ref{methods-1}), performing better than baselines on downstream tasks (Section~\ref{methods-2}). 
Our experiments on SemanticAN in Section~\ref{methods-1}, the ablation study model, revealed that SemanticGAN's performance is due to its ability to learn the joint image and segmentation distribution $p(x,y)$. 

The experiments on biased training in Section~\ref{training-bias} highlighted the model's reliance on the unlabelled dataset for generalisation to out-of-domain datasets.
In addition, counter-intuitively, the most significant relative drop in performance was observed in the labelled in-domain setting.
Further work should aim to understand the model's weaknesses to different biased training data.

Although SemanticGAN showed strong and consistent performance, the fully-supervised baseline consistently scored higher on segmentation metrics. 
An interesting follow-up study would be to test the hypothesis that the method's strengths outweigh its weaker performance in an active learning regime, whereby only the most informative examples are annotated. 
% However, we argue that although medical annotation is a scarce resource, collecting labelled datasets in the hundreds of examples is still an achievable target for any sensitive application. 
% In addition, Sup-UNet has 0.008\% of SemanticGAN's parameters, a ratio representative of the respective training time of the two models.

Our findings suggest that generative models are particularly well-suited when shape consistency is a critical desideratum. 
For other circumstances, the results reported in this work show that the method examined generalises well and does not exacerbate existing biases. 
However, SemanticGAN sometimes fails to perform better than a baseline that is easier to train and achieves excellent segmentation scores.
We conclude that generative approaches to medical image segmentation have potential and should be investigated in future work. 
They may become a viable alternative to discriminative models, in particular, due to their ability to incorporate information from unlabelled data.

\acks{MR is supported by UK Research and Innovation [UKRI Centre for Doctoral Training in AI for Healthcare grant number EP/S023283/1].}

\bibliography{Rosnati22}

\appendix
%\clearpage
\renewcommand\thefigure{A.\Roman{figure}}
\renewcommand\thetable{A.\Roman{table}}
%\section*{Appendix}
\setcounter{figure}{0}  
\setcounter{table}{0}  
% \FloatBarrier
% \usepackage{hhline}

\begin{table*}
\caption{Datasets used in our experiments}
\label{tab:datasets}
\centering
\setlength{\tabcolsep}{5pt}
\begin{tabular}{lcrrr}
\toprule
\textbf{Name} & \begin{tabular}[c]{@{}c@{}}\textbf{Segmentation }\\\textbf{labels}\end{tabular} & \textbf{\# scans} & \textbf{\# males} & \textbf{\# females}  \\ 
\midrule
ChestX-ray8   & No                                                                              & 108,000           & 60,999 (56\%)     & 47,001 (44\%)        \\
JSRT (train)  & Yes                                                                             & 350               & 171 (49\%)        & 179 (51\%)           \\
JSRT (test)   & Yes                                                                             & 81                & 35 (43\%)         & 46 (57\%)            \\
NIH*          & Yes                                                                             & 95                & 53 (56\%)         & 42 (44\%)            \\
Montgomery    & Yes                                                                             & 138               & 63 (46\%)         & 74 (54\%)            \\
CheXpert      & No                                                                              & 76,205            & 44,773 (59\%)     & 31,432 (41\%)        \\
\bottomrule
\end{tabular}\\
\footnotesize
*The NIH dataset is an annotated subset of ChestX-ray8.
\end{table*}

\section{Training details}
\label{sec:training-deets}
We resize the images to $256\times256$ resolution and normalise the pixel values to the range $[-1,1]$ for both images and segmentation maps to be compatible with StyleGAN's architecture.
In line with \cite{oh2020deep} prior to rescaling, we equalise the intensity histogram and apply a gamma correction of factor $0.5$.

For SemanticGAN, we use the same hyperparameters as \cite{li2021semantic}. We train the first step for 100k steps with a batch size of 8 on a single GPU. We choose the model at the step with the lowest FID score, and train the second step for 200k batches with a batch size of 4 (800k sample iterations).

For SemanticAN, we use the same hyperparameters as \cite{li2021semantic} but a smaller learning rate of $2\times 10^{-4}$. We train for 50k batches with a batch size of 32 (1,600k sample iterations) and choose the model with the highest validation DICE score.

For SupOnly, we use an Adam optimiser \citep{kingma2015adam} with learning rate of $10^{-5}$ and weight decay of $10^{-4}$. We train for 10k batches with a batch size of 32 (320k sample iterations) and choose the model with the highest validation DICE score.

Throughout the paper, any mention of statistical significance implies that we carried out a dependent t-test for paired variables and an independent Welch's t-test for independent variables, and the p-value was below 0.05.
\paragraph{Downstream task performance
for disease classification}
\label{sec:app-ex2}
For the classification model, we use a DenseNet \citep{huang2017densely} pre-trained on ImageNet from torchvision \citep{marcel2010torchvision}. We replace the last layer with a dense layer and randomised weights.
We use the CheXpert dataset, set the intensities of pixels outside the lung area to zero, and separate training and test sets. We experiment with rescaling the scans' intensities between 0 and 1, and find that the model performs better when the intensities are unnormalised -- between 0 and 255. 
We optimise the model with an Adam optimiser with learning rate of $10^{-3}$.

\section{Comparing U-Net and DeepLabV3 backbones}
\label{sec:Exp1-appendix}
This appendix chapter aims to compare the performance of DeepLabV3 and U-Net for fully supervised and semi-supervised segmentation. 
In Experiment 1 -- shown in Figure~\ref{exp1-app} -- when only using labelled images, DeepLabV3 outperforms both U-Net and all unsupervised methods in most testing circumstances. However, it is interesting to see that the adversarially trained U-Net performs comparably to the adversarially trained DeepLabV3 for the labelled in-domain data, and better for the unlabelled in-domain data and out-of-domain data. Similarly, in Experiment 2 -- shown in Figure~\ref{tab:unsup-appendix} -- SemanticAN trained with the U-Net backbone outperforms the DeepLabV3-based SemanticAN.
These results highlight that DeepLabV3 overfits to the labelled training data. 
\begin{table}[!t]
\centering
\caption{Diagnosis classification AUROC for images masked using different segmentation models on CheXpert (out-of-domain).}
\label{tab:unsup-appendix}
\setlength{\tabcolsep}{2pt}
\begin{tabular}{@{}lccc@{}} 
\toprule
%               & \begin{tabular}[c]{@{}c@{}}Lung\\ opacity\end{tabular} 
%               & \begin{tabular}[c]{@{}c@{}}Pleural\\effusion \end{tabular} & Oedema  \\
& Opacity & Effusion & Oedema\ \\
% \midrule
% \multicolumn{1}{c}{}& \multicolumn{3}{c}{CheXpert (out-of-domain)}           \\ 
% \cmidrule{2-4}

\midrule
%\begin{tabular}[c]{@{}c@{}}Unmasked\\ image\end{tabular} 

SupOnly UN       & 66.3                  & 80.0                      & 76.7            \\

SupOnly DL         & 67.5          & 82.0             & 78.0       \\
SemanticAN UN   & 67.1                  & 81.5                      & 77.3            \\

SemanticAN DL       & 66.8         & 78.8             & 77.3       \\

SemanticGAN    & \textbf{68.1} & \textbf{82.2}     & \textbf{78.4} \\
\midrule
%\begin{tabular}[c]{@{}c@{}}Unmasked\\ image\end{tabular} 
Unmasked     & \textbf{70.3}  & \textbf{85.3}    & \textbf{79.0}            \\
\bottomrule
\end{tabular}
\end{table}

\section{Downstream task subgroup performance}
\label{sec:Exp3-appendix}
\begin{figure*}[!t]
    \includegraphics[width=\textwidth]{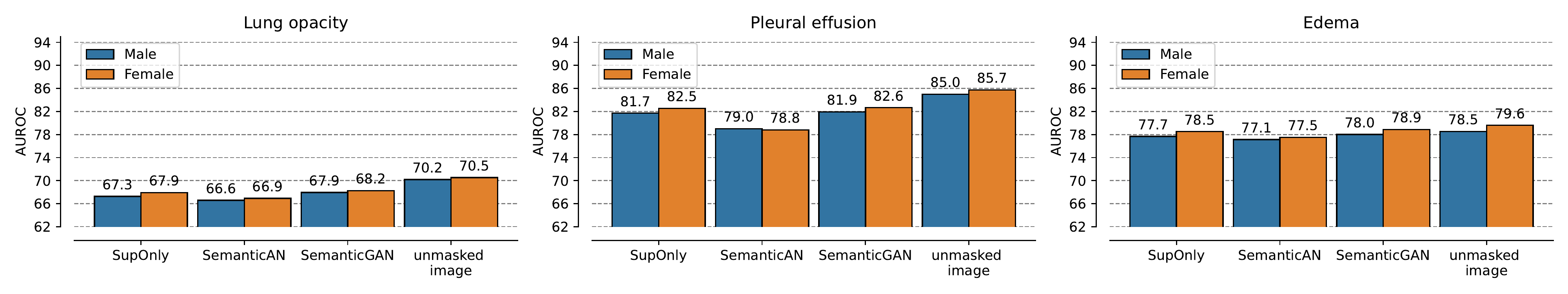}  
 \caption{Disease classification AUROC score for masked images stratified by biological sex.}
 \label{fig:unsup_sex}
\end{figure*}

Figure \ref{fig:unsup_sex} shows the stratification per sex of the AUROC for the diagnoses of lung opacity, pleural effusion and oedema using masked images derived in Section~\ref{methods-2}. 
The difference in the performance of models between males and females was not statistically significant, except for the unmasked image model, which performed better for females than males for oedema and pleural effusion. %(** for both).
Moreover, the difference in performance between models was in line with that observed in Section~\ref{methods-2}, where SemanticGAN outperformed SemanticAN and either performed similarly to SupOnly (lung opacity for females, pleural effusion, edema) or better (lung opacity for male), and the unmasked image model outperformed all other methods.
\begin{table*}[!t]
\centering
\caption{Models performance w.r.t.\ ground truth segmentations. Reported as mean $\pm$ standard deviation over the dataset. `UN' stands for U-Net and `DL' stands for DeepLabV3. }
\label{exp1-app}
\setlength{\tabcolsep}{3pt}
\begin{tabular}{@{}lccccc@{}}
\toprule
%            & \textbf{Dice}                    & \textbf{Precision}               & \textbf{Recall}                  & \begin{tabular}[c]{@{}c@{}}\textbf{Avg. surface }\\\textbf{distance}\end{tabular} & \begin{tabular}[c]{@{}c@{}}\textbf{Hausdorff }\\\textbf{distance}\end{tabular}  \\
            & Dice                    & Precision               & Recall                  & \begin{tabular}[c]{@{}c@{}}Avg. surface\\distance\end{tabular} & \begin{tabular}[c]{@{}c@{}}Hausdorff\\distance\end{tabular}  \\
% \midrule
\midrule
&\multicolumn{5}{c}{JSRT (labelled in-domain)}                                                                                                                                                                                                                                     \\ 
\cmidrule{2-6}
SupOnly UN    &     97.3 $\pm$ 1.1  & 98.4 $\pm$ 1.6          & 96.2 $\pm$ 1.7  & 1.4 $\pm$ 1.8 & 5.5 $\pm$ 9.3\\
SupOnly DL &\textbf{99.3 $\pm$ 0.4} & \textbf{99.1 $\pm$ 0.4} & \textbf{99.5 $\pm$ 0.5} & \textbf{0.3 $\pm$ 0.2} &\textbf{1.1 $\pm$ 0.4} \\
SemanticAN UN    & 93.2 $\pm$ 2.1  & 89.3 $\pm$ 3.5  & 97.6 $\pm$ 1.4 & 11.3 $\pm$ 4.5 & 54.4 $\pm$ 19.0\\
SemanticAN DL &91.2 $\pm$ 5.2 &93.3 $\pm$ 3.7 &89.8 $\pm$ 7.6 &3.7 $\pm$ 2.0 & 14.7 $\pm$ 7.1 \\
SemanticGAN &     98.2 $\pm$ 1.3  &        98.3 $\pm$ 1.5  &          98.1 $\pm$ 1.2 &         0.7 $\pm$ 0.6 &          2.5 $\pm$ 2.4\\ 
\midrule
&\multicolumn{5}{c}{NIH (unlabelled in-domain)}                                                                                                                                                                                                                                    \\ 
\cmidrule{2-6}
SupOnly UN        & 80.2 $\pm$ 15.9 & 91.0 $\pm$ 15.8         & 72.9 $\pm$ 17.2 & 8.6 $\pm$ 7.3 & 36.0 $\pm$ 24.4 \\
SupOnly DL& \textbf{90.3 $\pm$ 8.1} & \textbf{95.3 $\pm$ 4.6} & 86.5 $\pm$ 11.4 &          \textbf{4.8 $\pm$ 4.8} &25.5 $\pm$ 22.7  \\
SemanticAN UN   & 81.6 $\pm$ 9.6  & 76.3 $\pm$ 12.6 & 88.8 $\pm$ 7.7  & 22.1 $\pm$ 7.7 & 71.1 $\pm$ 17.6 \\
SemanticAN DL &72.4 $\pm$ 19.6 &75.4 $\pm$ 18.8 &71.2 $\pm$ 21.2 &10.9 $\pm$ 7.13 &35.5 $\pm$ 17.1 \\
SemanticGAN     & 90.1 $\pm$ 3.4 &          89.9 $\pm$ 6.5  &         \textbf{90.9 $\pm$ 5.0 } & 4.9 $\pm$ 3.3 & \textbf{18.2 $\pm$ 16.0}\\ 
\midrule
&\multicolumn{5}{c}{Montgomery (out-of-domain)}                                                                                                                                                                                                                                    \\ 
\cmidrule{2-6}
SupOnly UN    & 92.7 $\pm$ 7.9  & 96.6 $\pm$ 9.9  & 89.4 $\pm$ 6.9  & 3.5 $\pm$ 4.7 &13.7 $\pm$ 17.8 \\
SupOnly DL &\textbf{96.5 $\pm$ 2.3} &\textbf{98.8 $\pm$ 0.7} & 94.4 $\pm$ 4.0 &\textbf{1.3 $\pm$ 0.7} & \textbf{5.5 $\pm$ 7.2} \\
SemanticAN UN   & 90.3 $\pm$ 6.4  & 87.0 $\pm$ 8.9  & \textbf{94.5 $\pm$ 4.0}  & 17.6 $\pm$ 8.5 & 72.3 $\pm$ 27.2 \\
SemanticAN DL &57.9 $\pm$ 33.4 &62.4 $\pm$ 31.0 &55.5 $\pm$ 35.1 &19.3 $\pm$ 15.1 &54.8 $\pm$ 38.9 \\
SemanticGAN &     91.3 $\pm$ 5.6  &         92.6 $\pm$ 7.9  &         90.4 $\pm$ 4.2  &         8.8 $\pm$ 10.5 & 34.2 $\pm$ 39.1        \\
\bottomrule
\end{tabular}
\end{table*}

\begin{table*}[!ht]
\centering
\caption{Detail of training population for Experiment~\ref{subgroup-perf}, where ``all" refers to both male and female samples being included.}
\label{tab:detail-ex-bias}
\setlength{\tabcolsep}{8pt}
\begin{tabular}{lcccc}
\toprule
\multirow{2}{*}{Experiment name}   & \multicolumn{2}{c}{$G$ training}                                       & \multicolumn{2}{c}{$E$ training}                                         \\
\cmidrule{2-5}
                                   & $\mathcal{D}_u$ & $\mathcal{D}_l$ & $\mathcal{D}_u$ & $\mathcal{D}_l$  \\ 
\midrule
Control                   & all                                 & all                                 & all                                 & all                                  \\
Full bias                 & males                               & males                               & males                               & males                                \\
Biased generator $G$          & males                               & males                               & all                                 & all                                  \\
Biased encoder $E$            & all                                 & all                                 & males                               & males                                \\
Biased labelled dataset $\mathcal{D}_l$   & all                                 & males                               & all                                 & males                                \\
Biased unlabelled dataset $\mathcal{D}_u$ & males                               & all                                 & males                               & all                                 \\
\bottomrule
\end{tabular}
\end{table*}
\begin{table}[!ht]
\centering
\caption{Stratified models performance for females and males, quantified by Dice similarity coefficient. `ID' stands for in-domain, `OOD' stands for out-of-domain.}
\label{fig:sup-sex-numbers}
\setlength{\tabcolsep}{1pt}
\begin{tabular}{@{}lcc@{}}
\toprule
& Female & Male    \\
\midrule
&\multicolumn{2}{c}{JSRT (lab.\ ID)} \\
\cmidrule{2-3}
SupOnly UN  & 97.0~$\pm$~1.4 & 97.5~$\pm$~0.5  \\
SupOnly DL & \textbf{99.3 $\pm$ 0.4} & \textbf{99.3 $\pm$ 0.4} \\
SemanticAN UN    & 92.5~$\pm$~2.6 & 93.9~$\pm$ 1.0   \\
SemanticAN DL & 89.9 $\pm$ 7.1& 92.3 $\pm$ 1.9\\ 
SemanticGAN & 98.2 $\pm$ 1.1 & 98.2~$\pm$ 1.5   \\
\midrule
&\multicolumn{2}{c}{NIH (unlab.\ ID)} \\
\cmidrule{2-3}
SupOnly UN   &  77.3~$\pm$ 18.1 & 82.4~$\pm$ 13.8 \\
SupOnly DL &    88.9 $\pm$ 9.6 & \textbf{91.4 $\pm$ 6.6}\\
SemanticAN UN & 79.0~$\pm$~9.8  & 83.6~$\pm$ 9.0 \\
SemanticAN DL & 67.5 $\pm$ 22.6& 76.3 $\pm$ 16.0\\ 
SemanticGAN &   90.5~$\pm$ 2.8  & 89.8~$\pm$~3.9 \\
\midrule
&\multicolumn{2}{c}{Montgomery (OOD)} \\
\cmidrule{2-3}
SupOnly UN    & 91.8~$\pm$~8.4 & 94.2~$\pm$~5.3 \\
SupOnly DL    & 96.5 $\pm$ 2.3& \textbf{96.5 $\pm$ 2.2}\\
SemanticAN UN & 89.5~$\pm$~5.8 & 91.7~$\pm$~6.2 \\
SemanticAN DL & 44.1 $\pm$ 31.2& 74.8 $\pm$ 27.7\\ 
SemanticGAN & 90.6~$\pm$ 4.5 & 92.3~$\pm$~6.4 \\
\bottomrule
\end{tabular}
\end{table}
\begin{table}[!ht]
\caption{Biological sex bias impact on model performance}
    \label{fig:sex_bias_numerical}

\centering
\setlength{\tabcolsep}{2pt}
\begin{tabular}{@{}lcc@{}}
\toprule
& Female& Male\\
\midrule
&\multicolumn{2}{c}{JSRT (lab.\ in-domain)} \\
\cmidrule{2-3}
Control                & 98.2~$\pm$~1.1 & 98.2~$\pm$~1.5  \\
Full bias              & 93.8~$\pm$~2.7 & 98.3~$\pm$~1.6  \\
Biased $G$             & 98.6~$\pm$~0.9 & 98.4~$\pm$~1.4   \\
Biased $E$             & 94.3~$\pm$~1.7 & 98.5~$\pm$~1.6   \\
Biased $\mathcal{D}_l$ & 94.8~$\pm$~1.3 & 98.5~$\pm$~1.6  \\
Biased $\mathcal{D}_u$ & 98.4~$\pm$~1.1 & 98.2~$\pm$~1.7   \\
\midrule
&\multicolumn{2}{c}{NIH (unlab.\ in-domain)} \\
\cmidrule{2-3}
Control                & 90.6~$\pm$~2.8 & 89.8~$\pm$~3.9 \\
Full bias              & 89.7~$\pm$~5.1 & 91.0~$\pm$~3.5  \\
Biased $G$             & 92.3~$\pm$~2.8 & 92.3~$\pm$~3.0 \\
Biased $E$             & 88.3~$\pm$~4.2 & 90.0~$\pm$~3.6   \\
Biased $\mathcal{D}_l$ & 90.6~$\pm$~3.5 & 91.2~$\pm$~2.9   \\
Biased $\mathcal{D}_u$ & 88.5~$\pm$~4.6 & 89.9~$\pm$~2.8   \\
\midrule
&\multicolumn{2}{c}{Montgomery (out-of-domain)} \\
\cmidrule{2-3}
Control                & 90.6~$\pm$~4.5 & 92.3~$\pm$~6.5 \\
Full bias              & 88.4~$\pm$~5.3 & 92.5~$\pm$~5.2   \\
Biased $G$             & 92.4~$\pm$~3.3 & 94.0~$\pm$~3.5 \\
Biased $E$             & 90.3~$\pm$~3.9 & 93.0~$\pm$~3.8   \\
Biased $\mathcal{D}_l$ & 91.2~$\pm$~3.2 & 93.2~$\pm$~3.7   \\
Biased $\mathcal{D}_u$ & 88.8~$\pm$~4.0 & 92.0~$\pm$~5.2  \\
\bottomrule
\end{tabular}
\end{table}
\section{Consistency over continuous image changes}
\label{PCA}
As a final experiment, we aim to gain a deeper understanding of SemanticGAN from a qualitative perspective by applying it to PCA-generated neighbouring samples, observing both the image reconstruction and the segmentation. We interpolate scans through PCA instead of well-known generative adversarial interpolation methods to avoid the circularity of using a tool to evaluate itself.
We proceed by training a PCA model using 3,000 random images from CheXpert balanced across biological sex and ethnicity. We extract the mean PCA image and four neighbouring images at 1 and 2 standard deviations in each direction for each of the first three principal modes.
We then use SemanticGAN to predict the lung segmentation and qualitatively examine the gradual change in both the image reconstructions and label maps.

\begin{figure*}[!t]
    \centering
    \includegraphics[width=\textwidth]{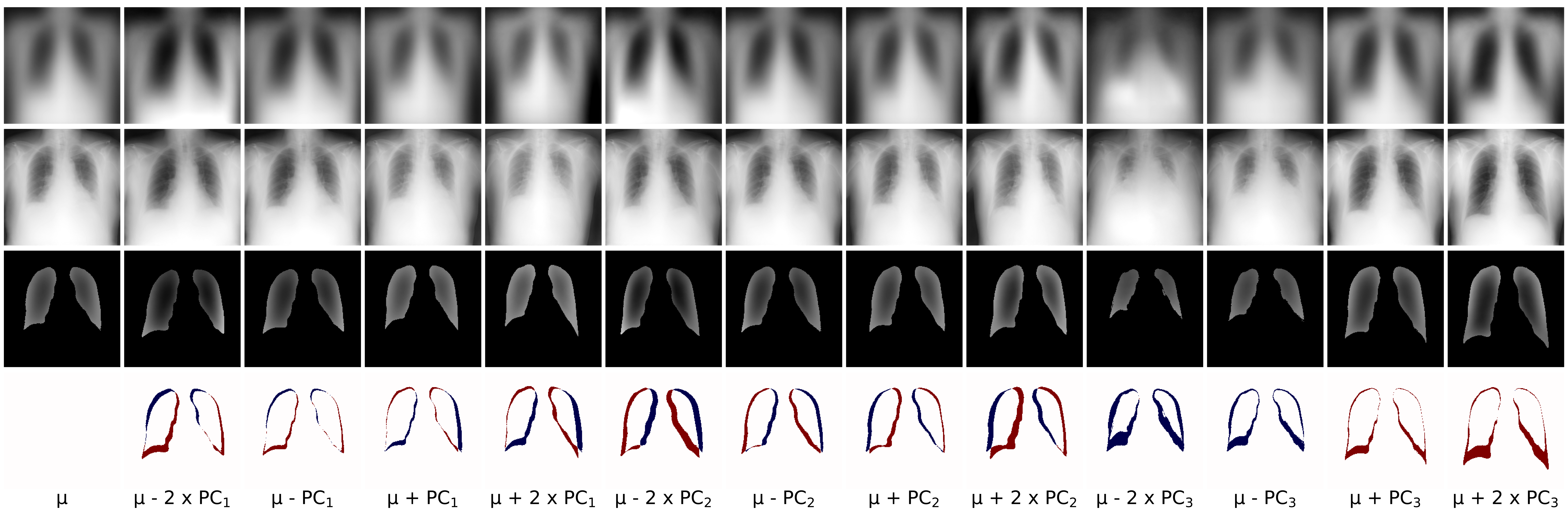}
     \caption{CheXpert dataset PCA, with generative model reconstruction and segmentation. The first row shows the principal component images, the second row shows SemanticGAN's reconstruction, the third row show its segmentation, and the fourth row shows the difference in segmentation with the mean PCA image.}
    \label{fig:PCA}
\end{figure*}
Figure \ref{fig:PCA} shows SemanticGAN's image reconstruction and segmentation for different PCA components. 
The first principal component (columns 2-5) reflects the relative closeness of the patient to the X-ray source. In the left-most image, the patient occupies a more significant percentage of space: the space between the image's border and the patient is small, and the lung fields appear larger. Conversely, in the right-most PCA component, the space between the patient and the image borders is more significant, and the patient's lung areas are smaller. SemanticGAN's segmentations reflect the patient closeness, where the left-most segmentation is 9\% larger than the mean, and the right-most segmentation is 10\% smaller than the mean. 
For the second component (columns 6-9), the main change is in the patient positioning, from left to right of the field of view, as can be seen from the image margins. Accordingly, the left-most image segmentation is further left than the mean one, and the right-most image segmentation is further right than the mean one. 
Finally, for the third component (columns 10-13), the size of the lung area increases from left to right. The segmentation grows accordingly: the left-most segmentation is 49\% smaller than the mean one, and the right-most segmentation is 39\% larger than the mean one.
Interestingly, the model's image reconstruction appeared more realistic than the PCA image, providing a good sanity check of the mechanisms underpinning the model's segmentation production. 
We found that images which were close in PCA space had similar segmentations. 
Although the shape of the lung area could not be distinguished, SemanticGAN generated segmentations with consistent shapes but varying sizes.
An intuitive explanation for this behaviour is that SemanticGAN learns a continuous template of what the lung segmentation resembles and subsequently adapts it to the given image.

% Daniel Coelho de Castro: You may want to refer to the size of the lung fields (i.e. visible lung shadow on the image), and relate that to e.g. inspiration.

\FloatBarrier
\begin{figure*}[!ht]\label{fig:ds-thumb}
    \centering
    \subfigure[ChestX-ray8][b]{
    \includegraphics[width=\textwidth]{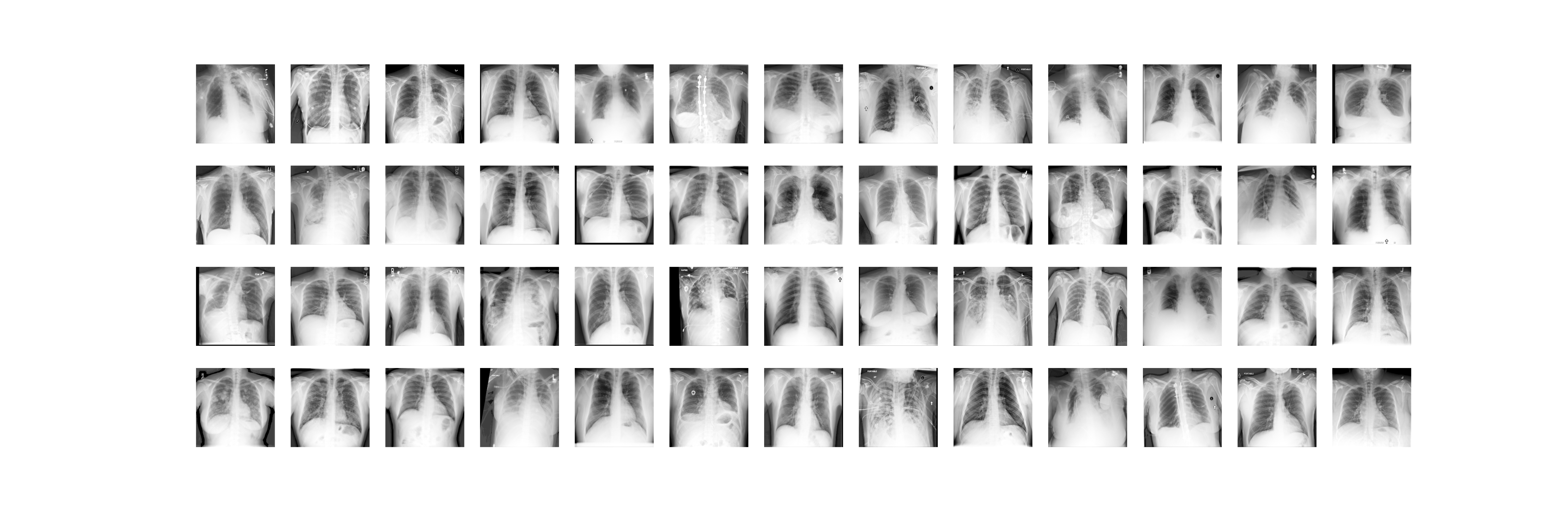}
}
 
    \subfigure[JSRT][b]{
    \centering
    \includegraphics[width=\textwidth]{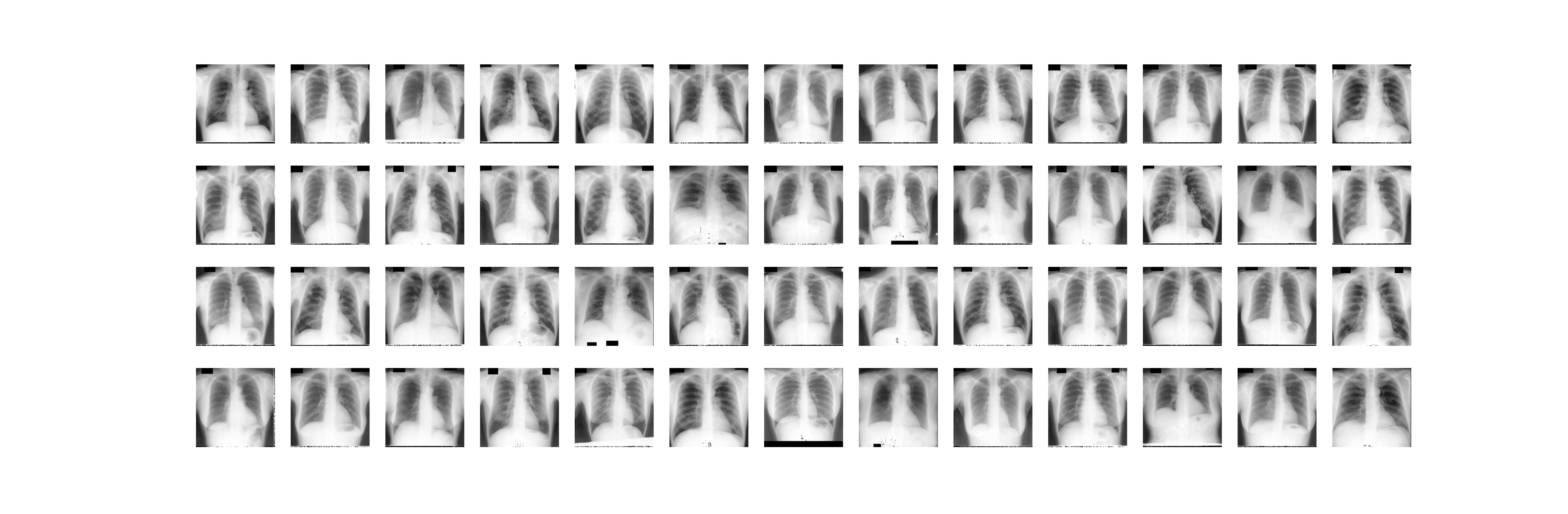}
    }

    \subfigure[NIH][b]{
    \centering
    \includegraphics[width=\textwidth]{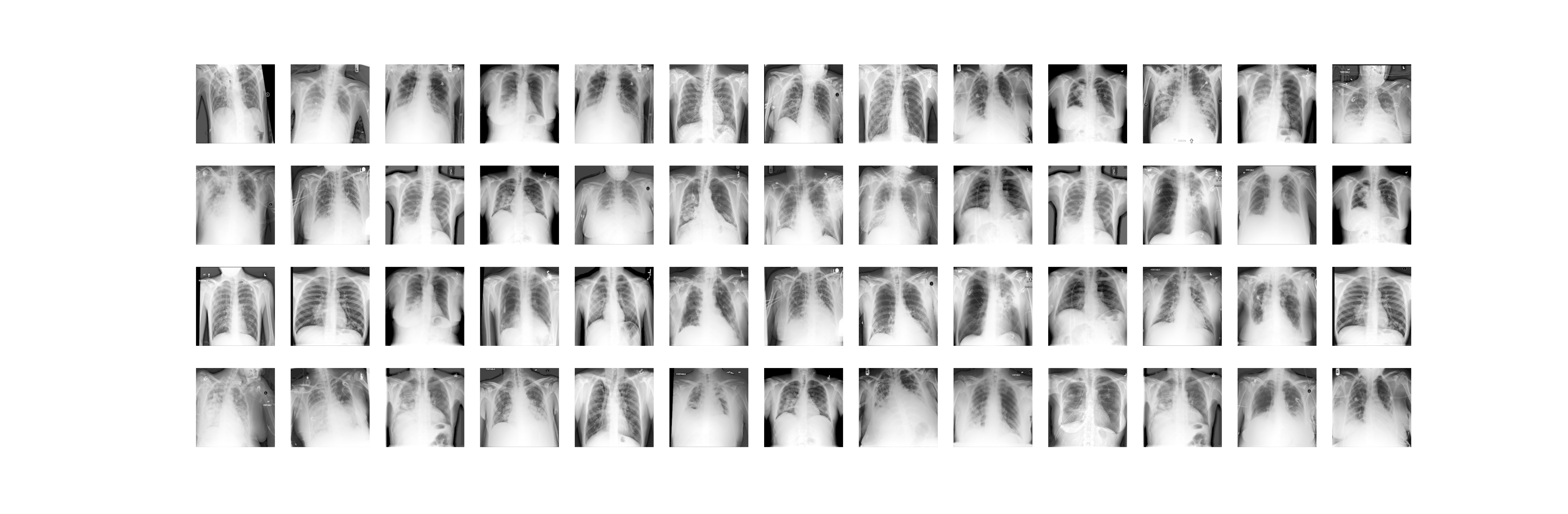}}
     \caption{Datasets thumbnails}
\end{figure*}
 \begin{figure*}[!ht]      
    \subfigure[Montgomery][b]{
    \centering
    \includegraphics[width=\textwidth]{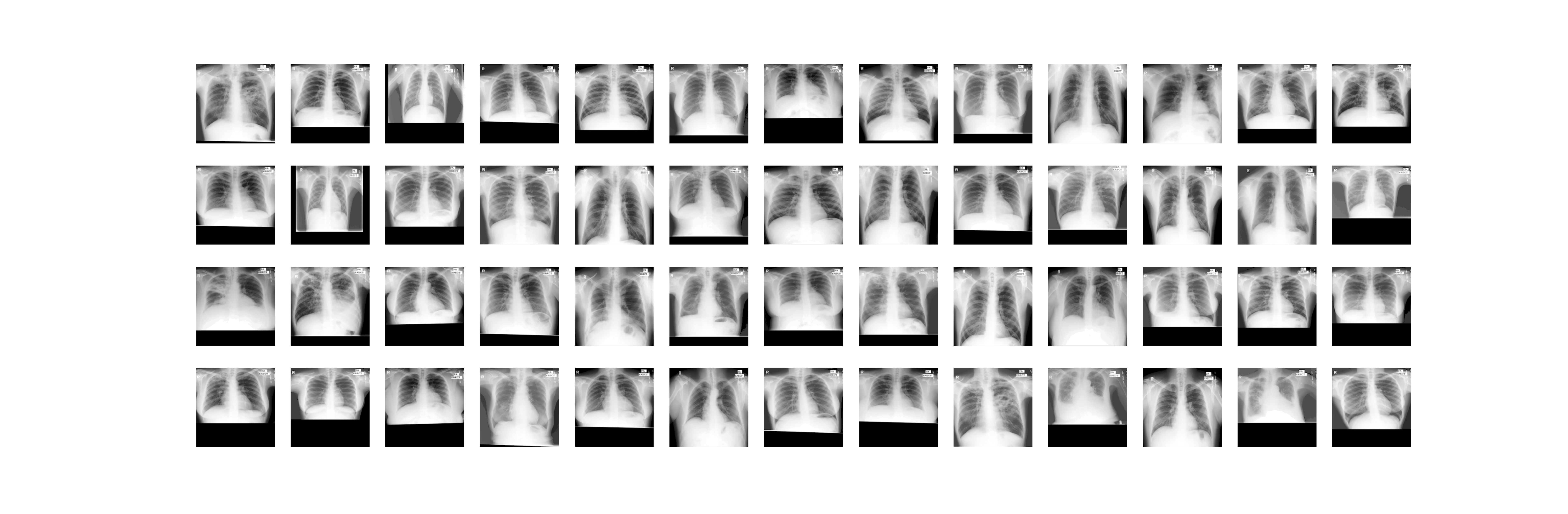}}
    
    \subfigure[CheXpert][b]{
    \centering
    \includegraphics[width=\textwidth]{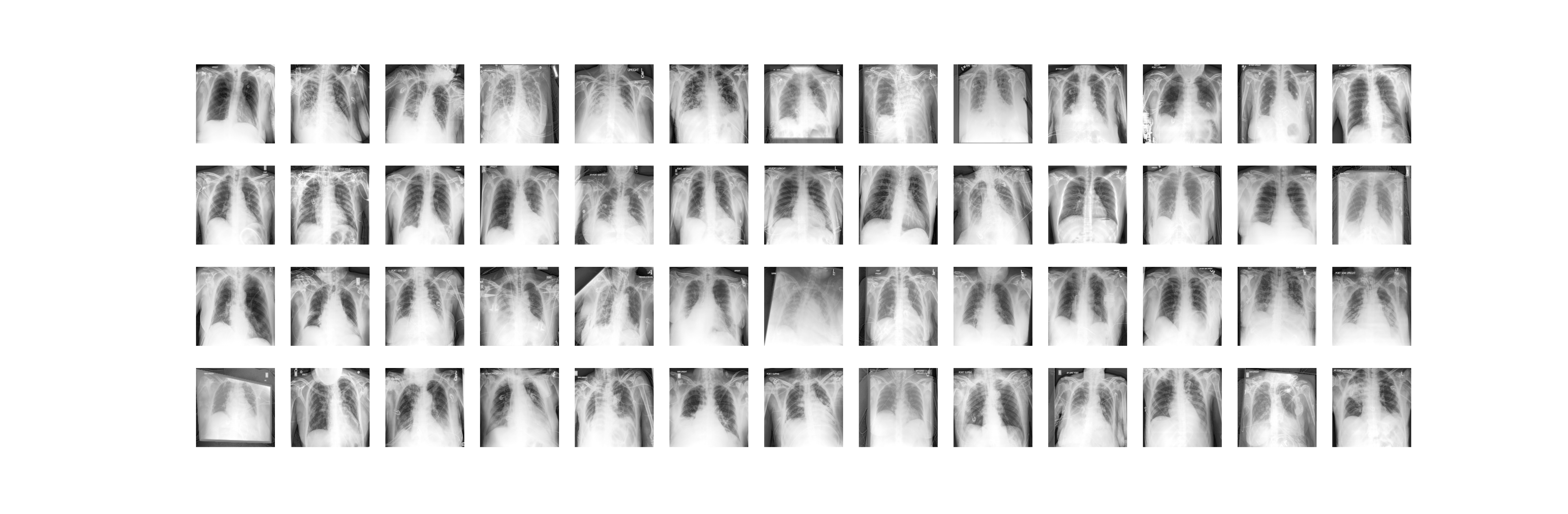}}
     \caption{Datasets thumbnails - continued}
    \label{fig:ds-thumb-2}
\end{figure*}

\end{document}